\documentclass[runningheads]{llncs}
\usepackage{graphicx}
\usepackage{amsmath}
\usepackage{color}

\usepackage{fancyhdr}
\fancypagestyle{specialfooter}{%
  \fancyhf{}
  
  \fancyfoot[L]{\scriptsize{$^2$Work done while at IIIT Sri City}}
  \fancyhead[L]{\scriptsize{Accepted in Fourth Workshop on Computer Vision Applications (WCVA) at ICVGIP 2021}}
}

\begin{document}

\title{Semantic Map Injected GAN Training for Image-to-Image Translation}

\author{Balaram Singh Kshatriya$^1$, Shiv Ram Dubey$^2$, Himangshu Sarma$^1$, Kunal Chaudhary$^3$, Meva Ram Gurjar$^3$, Rahul Rai$^3$, Sunny Manchanda$^3$}

\authorrunning{B.S. Kshatriya et al.}

\institute{Computer Vision Group, Indian Institute of Information Technology, Sri City, Chittoor, India \and Computer Vision and Biometrics Lab, Indian Institute of Information Technology, Allahabad, India 
\and DRDO Young Scientist Laboratory - Artificial Intelligence (DYSL-AI), Bangalore, India
\\
{\tt\small balaramsingh.k18@iiits.in, srdubey@iiita.ac.in, himangshu.sarma@iiits.in, \{kunal, mrgurjar, rahulrai, sunny\}@dysl-ai.drdo.in}
}

\maketitle
\thispagestyle{specialfooter}

\begin{abstract}
Image-to-image translation is the recent trend to transform images from one domain to another domain using generative adversarial network (GAN). The existing GAN models perform the training by only utilizing the input and output modalities of transformation. In this paper, we perform the semantic injected training of GAN models. Specifically, we train with original input and output modalities and inject a few epochs of training for translation from input to semantic map. Lets refer the original training as the training for the translation of input image into target domain. The injection of semantic training in the original training improves the generalization capability of the trained GAN model. Moreover, it also preserves the categorical information in a better way in the generated image. The semantic map is only utilized at the training time and is not required at the test time. The experiments are performed using state-of-the-art GAN models over CityScapes and RGB-NIR stereo datasets. We observe the improved performance in terms of the SSIM, FID and KID scores after injecting semantic training as compared to original training.
\end{abstract}

\section{Introduction}
Deep learning has shown very promising growth in the past decade \cite{lecun2015deep}, \cite{dubey2021decade}. Convolutional neural network (CNN) has led to huge progress in deep learning \cite{alexnet}. CNNs have been heavily used to deal with the image data for different applications such as image recognition \cite{alexnet}, \cite{resnet}, face recognition \cite{srivastava2019performance}, \cite{liu2017sphereface}, medical analysis \cite{wang2017chestx}, \cite{basha2018rccnet},
depth estimation \cite{garg2016unsupervised}, \cite{repala2019dual}, and many more.

Generative adversarial network (GAN) is introduced by Goodfellow et al. \cite{goodfellow2014generative} for generating the new samples for any given data distribution. GAN consists of two networks, namely generator and discriminator. The generator network generates the samples from the random noise vector. The discriminator network facilitates the training of the generator network by classifying the real and generated samples. The training of both generator and discriminator networks is performed in a min-max optimization fashion. The generator network tries to generate realistic samples using the training distribution such that it can fool the discriminator network. However, the discriminator network tries not to get fooled by the generator network by classifying the generated sample into a fake category.
GAN has been utilized for different applications such as data generation \cite{mao2017least}, \cite{gulrajani2017improved}, medical image synthesis \cite{han2018gan}, \cite{frid2018gan}, minority class oversampling \cite{mullick2019generative}, \cite{roy2021generative}, image hashing \cite{cao2018hashgan} among others.

\begin{figure}[!t]
    \centering
    \includegraphics[width=\textwidth]{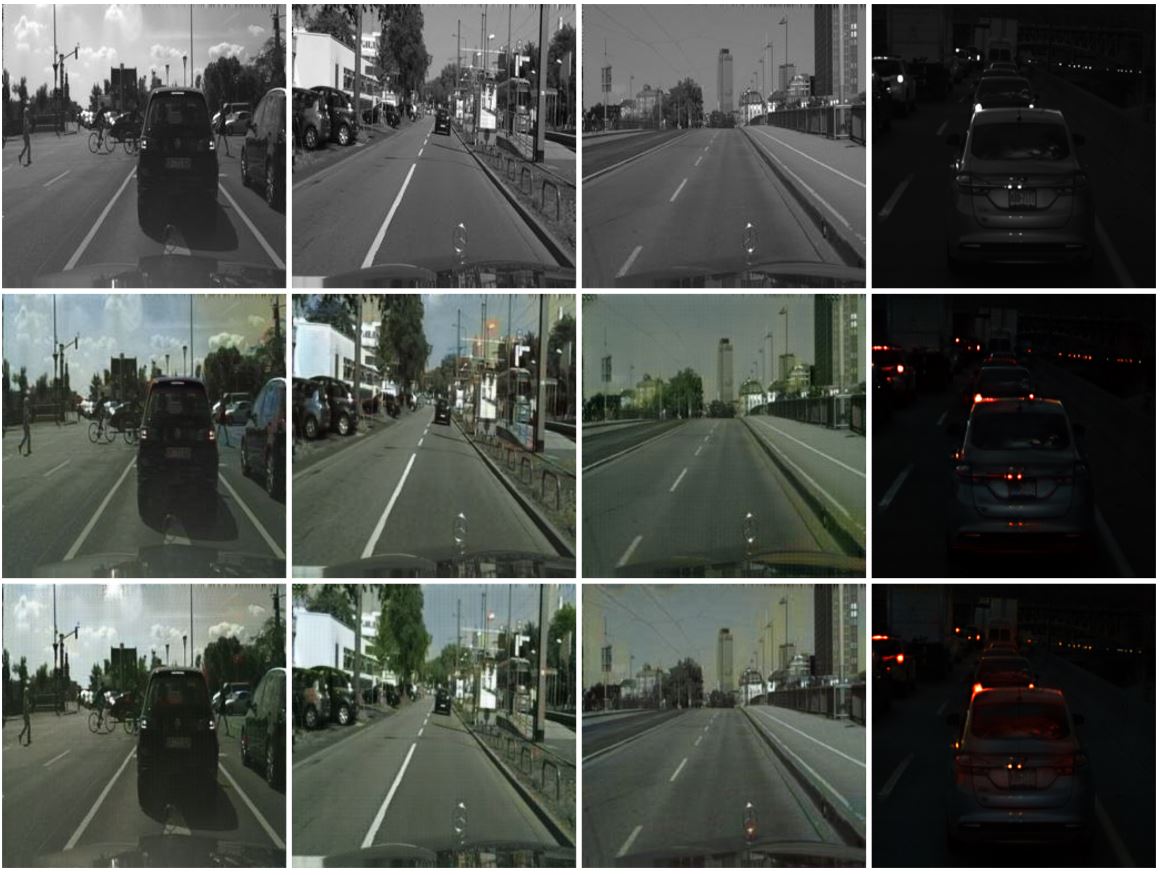}
    \caption{\small{Outputs of image to image translation from monochrome to RGB. $1^{st}$ \textit{row}: the input images, $2^{nd}$ \textit{row}: the generated images using CycleGAN model \cite{cyclegan}, and $3^{rd}$ \textit{row}: the generated images using CUT model \cite{cut}.} }
    \label{fig:motivation}
\end{figure}

GAN has also shown very appealing performance for image to image translation \cite{pix2pix}, \cite{cyclegan} \cite{csgan}, \cite{pcsgan}, \cite{cdgan}, \cite{cut}. Broadly, it can be categorized into two parts, viz. paired and unpaired image to image translation. In case of the paired scenario, the source and target images are paired. The Pix2Pix model \cite{pix2pix} using conditional GAN relies on the paired data. The paired image to image translation requires the paired source and target domain images which is very laborious and infeasible to collect in many real applications. 
The CycleGAN model \cite{cyclegan} is designed for unpaired image to image translation. Two generator and two discriminator networks are utilized by CycleGAN. Forward generator transforms from source domain to target while backward generator transforms from target domain to source domain. Thus, an image is transformed from source domain to target domain using forward generator and cycled back to source domain using backward generator. A cycle consistency loss is utilized between the original image and the cycled image. A cyclic synthesized loss is included in cyclic synthesized GAN (CSGAN) model \cite{csgan} between the cycled image in a domain (A $\rightarrow$ B' $\rightarrow$ A'') and synthesized image in the same domain (B $\rightarrow$ A'). The perceptual loss is used between the original image in a domain and the synthesized image in the same domain by the perceptual cyclic synthesized GAN (PCSGAN) model \cite{pcsgan}. The mapping from source domain to target domain is constrained in both CSGAN and PCSGAN which is suitable for paired image translation. The CDGAN model \cite{cdgan} improves the CycleGAN framework for unpaired translation by adding the discriminator networks for cycled images. In order to synthesize the objects in a distinguishable fashion, contrastive learning is utilized in contrastive unpaired translation (CUT) model \cite{cut}. The CUT model considers a patch in the generated image and finds the positive and negative pairs of patches from the source domain. It utilizes the contrastive loss on the features of positive and negative pairs to ensure that the patches should be distinguishable from the negative patches which inherently improves the visible quality of the synthesized images. Other GAN models for image to image translation includes DualGAN \cite{yi2017dualgan}, AttentionGAN \cite{mejjati2018unsupervised}, CouncilGAN \cite{nizan2020breaking}, Multi-Scale Gradient based U-Net (MSG U-Net) \cite{msgunet}, etc. The GAN based image to image translation has been also utilized for image colorrization \cite{lee2020reference}, \cite{wu2021towards}, \cite{zhang2016colorful}.

\begin{figure}[!t]
    \centering
    \includegraphics[width=\columnwidth]{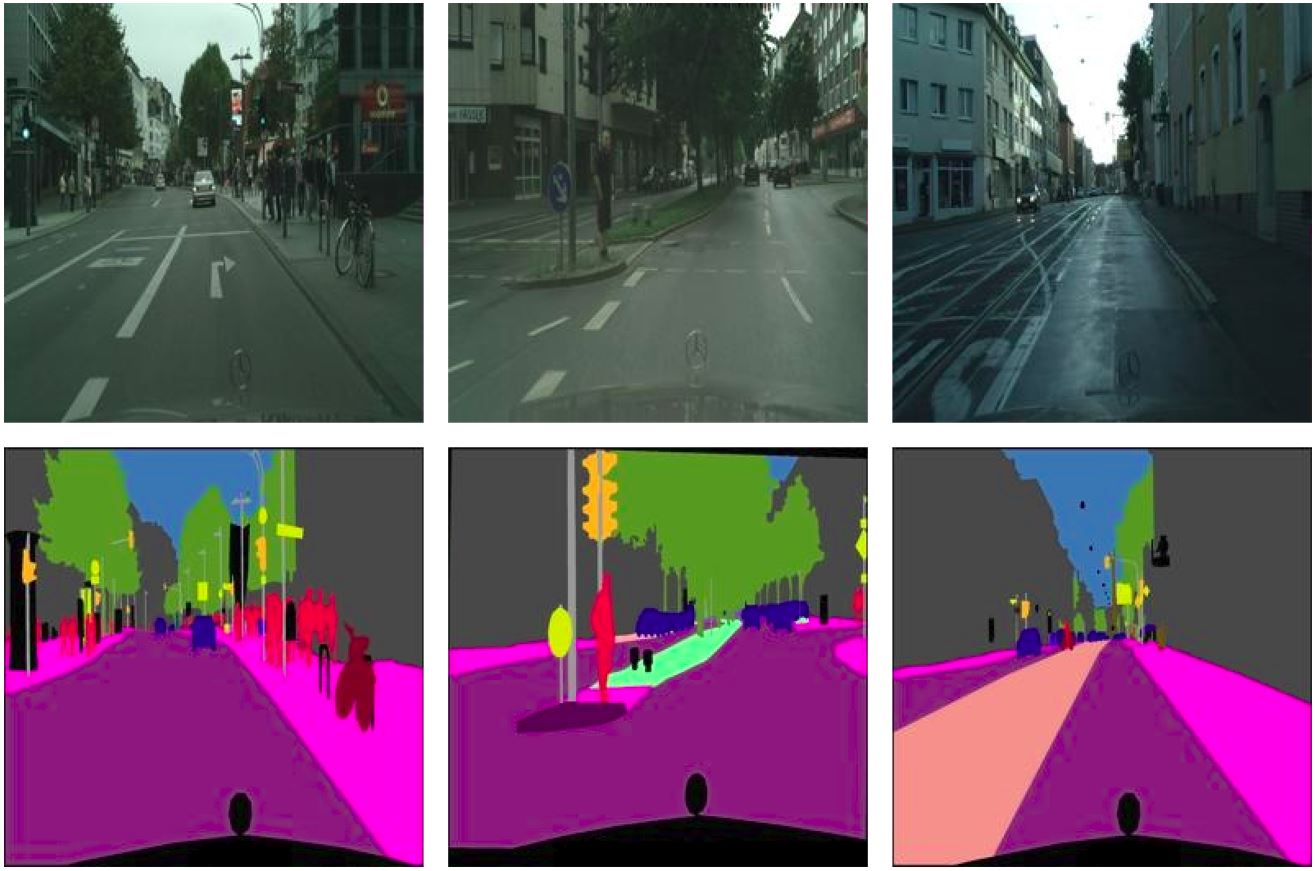}
    \caption{Samples of semantic maps in the $2^{nd}$ row corresponding to RGB images in $1^{st}$ row.}
    \label{fig:Semantic_image}
\end{figure}

There have been many recent works in the image to image translation which utilize semantic image along with source domain and the target domain in the translation. Sem-GAN \cite{cherian2019sem} performs image to image translation by utlizing the semantic information with the help of cycleGAN framework. SemGAN includes an additional loss named consistency constraints along with existing GAN loss and cycle consistency loss. Exemplar guided unsupervised image to image translation with semantic consistency \cite{ma2018exemplar} uses feature masks to avoid the semantic inconsistency. This allows to transfer style information of the target image. Segmentation guided image to image translation with adversarial networks \cite{jiang2019segmentation} is designed to impose semantic information on the generated images. An additional network named segmentor is incorporated on the the existing architecture which provides spatial guidance to the generator network. Example guided style consistent image synthesis from semantic learning \cite{wang2019example} is constructed by including a consistency discriminator which functions along with the existing generator and discriminator network to enforce style consistency on the given source images. All these models perform image to image translation by including semantic images with the help of additional network(s) to utilize the semantic information. However, the proposed network utilize the semantic information in the training schedule without any additional network.

The existing image to image translation methods using GAN are unable to take care of the semantic translation in terms of the structure and color of the object categories. Basically, it is often found that these GAN models of image to image translation overfit a particular color or object characteristic across the training dataset, resulting in incoherent translation of the given input image. Specifically, these networks learn the mapping from one domain to other, but fail to learn about the regions of each category in the image, which may lead to poor generalization over the test data.
It can be visualized in Fig. \ref{fig:motivation} which illustrates the generated images using CycleGAN \cite{cyclegan} in $2^{nd}$ row and CUT \cite{cut} models in $3^{rd}$ row for the sample input images shown in $1^{st}$ row. It can be noticed that the green color is overfitted and also some portions of images are still in monochrome (i.e., car in example 2 and buildings in example 3). These examples clearly indicate the limitation of the existing models, which lead to improper image translation.

Motivated by the limitations of the existing GAN models, we propose to utilize semantic map information while training. It can be seen in Fig. \ref{fig:Semantic_image} that the semantic map can better provide the object specific information.
Given the fact that the semantic maps might not be available at test time, the proposed approach uses the semantic map in the form of injected training such that it is not needed at the test time. We observe the improved performance of the proposed semantic injected GAN training approach.
Following are the major contributions of this work:
\begin{itemize}
    \item The proposed approach utilizes the semantic map to learn better category specific features.
    \item The proposed approach injects the transformation of input image to segmentation map during the training of transformation of input image to target image. 
    \item The proposed semantic injection improves the generalization ability of image translation.
    \item The use of semantic maps is required only at the training time not at the test time. 
    \item The performance of the proposed semantic injected training is tested using CycleGAN and CUT models over two benchmark datasets for image to image translation shows improved performance.
\end{itemize}

Remaining paper is organized as follows: Section 2 presents the proposed method; Section 3 summarizes the experimental settings, Section 4 illustrates the results and analysis; and Section 5 concludes the paper.

\section{Proposed Semantic Map Injected GAN Training}
\label{proposedmethod}

\begin{figure}[!t]
    \centering
    \includegraphics[width=0.45\textwidth]{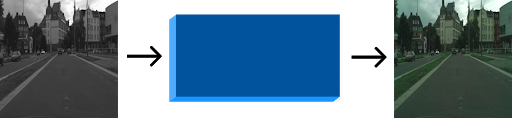}
    \hspace{0.05\textwidth}
    \includegraphics[width=0.45\textwidth]{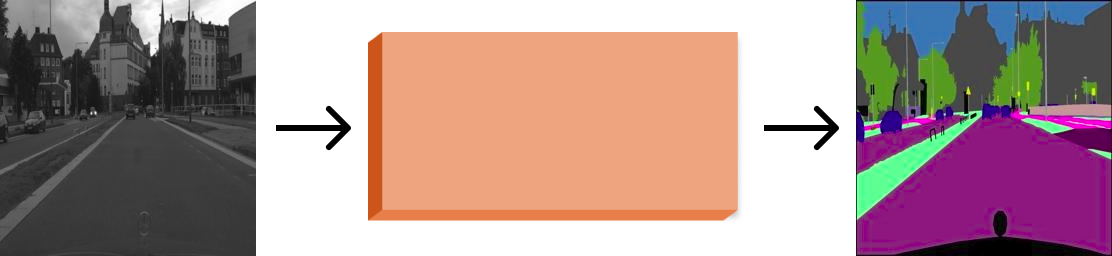}
    \caption{Representation of image to image translation from (left) input domain to target domain and (right) input domain to semantic domain.} 
    \label{fig:alternate}
\end{figure}

\begin{figure}[!t]
    \includegraphics[width=0.225\textwidth]{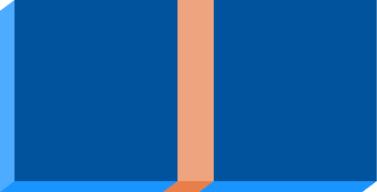}
    \hspace{0.01\textwidth}
    \includegraphics[width=0.225\textwidth]{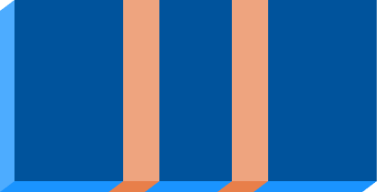}
    \hspace{0.01\textwidth}
    \includegraphics[width=0.225\textwidth]{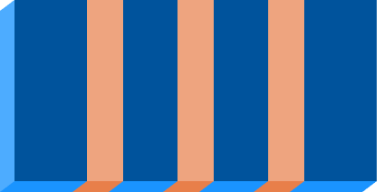}
    \hspace{0.01\textwidth}
    \includegraphics[width=0.225\textwidth]{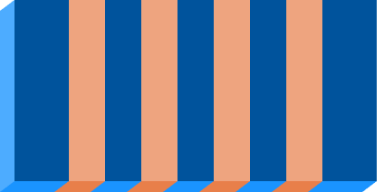}
    \caption{Representation of the ratio between the original training and semantic map training (i.e., Original:Semantic). In order (left to right): 90:10, 80:20, 70:30, and 60:40.}
    \label{fig:semantic_injected_training}
\end{figure}

\begin{figure*}[!t]
    \centering
    \includegraphics[width=0.9\textwidth]{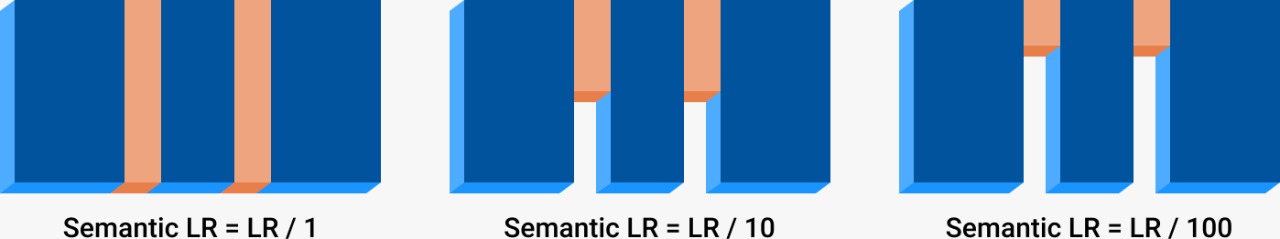}
    \caption{Representation of learning rate (LR) setting for 80:20 training model. LR Setting 1 (\textit{left}): semantic training LR is same as original training LR, LR Setting 10 (\textit{middle}): semantic training LR is one tenth of original training LR, and LR Setting 100 (\textit{right}): semantic training LR is one hundredth of original training LR.}
    \label{fig:lr}
\end{figure*}

The semantic map can be utilized for training of the GAN models in multiple ways. One of the obvious ways is to simply concatenate the semantic image to the given input image and train the model. This type of training might be effective, however it leads to the burden of having the semantic map also at the test time, which limits its uses in most real world applications. Thus, in this paper, we propose a novel way of utilizing the semantic map information at the training time by injecting the semantic map training. Specifically, we alternate the training of GAN models on target images and given semantic images while keeping the same input images in the source domain.

The visual representation of this procedure can be perceived in Fig. \ref{fig:alternate}.
Let us consider that the blue block represents the training of the GAN models for image to image translation from input domain to target domain and the red block represents the training the GAN models for image to image translation from input domain to semantic domain.
We propose to inject the semantic training in original training in an interleaved fashion as described in Fig. \ref{fig:semantic_injected_training}. The injection of semantic training is carried out in the chunks of $y$ number of epochs of training. Thus, based on the number of chunks of semantic training, the ratio between the number of epochs for original training and semantic training vary. If the total number of epochs for training is denoted by $n$ and the number of semantic training injection is $s$, then the ratio between the number of epochs for original training and semantic training (original:semantic) can be given as follows:
\begin{equation}
    original:semantic = (n - s \times y) : (s \times y)
\end{equation}
where $(s \times y)$ is the number of epochs of semantic training and $(n - s \times y)$ is the number of epochs of original training. The original:semantic training ratio is one of the hyperparameters in the proposed model.
Note that as the semantic training is interleaved with original training, we consider five training models (with total number of training epochs $n=100$) with different original:semantic training ratio. 
Fig. \ref{fig:semantic_injected_training} shows the training strategies with different values of original:semantic ratio. Following are the different training schedules used in this paper in terms of the sequence of number of epochs of original and semantic training: 
\begin{itemize}
\item 100:0 model (\textcolor{blue}{100})
\item 90:10 model (\textcolor{blue}{45}, \textcolor{red}{\textbf{10}}, \textcolor{blue}{45})
\item 80:20 model (\textcolor{blue}{30}, \textcolor{red}{\textbf{10}}, \textcolor{blue}{20}, \textcolor{red}{\textbf{10}}, \textcolor{blue}{30})
\item 70:30 model (\textcolor{blue}{20}, \textcolor{red}{\textbf{10}}, \textcolor{blue}{15}, \textcolor{red}{\textbf{10}}, \textcolor{blue}{15}, \textcolor{red}{\textbf{10}}, \textcolor{blue}{20})
\item 60:40 model (\textcolor{blue}{15}, \textcolor{red}{\textbf{10}}, \textcolor{blue}{10}, \textcolor{red}{\textbf{10}}, \textcolor{blue}{10}, \textcolor{red}{\textbf{10}}, \textcolor{blue}{10}, \textcolor{red}{\textbf{10}}, \textcolor{blue}{15})
\end{itemize}
where the epochs in \textcolor{blue}{blue} color represents the number of training epochs for image translation to target image and \textcolor{red}{red} color represents the number of training epochs for image translation to semantic map. The 100:0 model is without semantic training and considered for the comparison purpose to show the impact of semantic training injection. The number of semantic training epochs in each chunk (i.e., $y$) is 10 in the experiments. The number of semantic training chunks (i.e., $s$) is 0, 1, 2, 3 and 4 in 100:0, 90:10, 80:20, 70:30 and 60:40 training settings, respectively. Note that these training schedules inject the semantic training in the training schedule in an interleaved manner to retain the symmetry and to avoid the biasness towards the semantic translation. Our hypothesis is that the injection of semantic training in this fashion does not deviate from the original training, rather serves as regularizer to avoid the overfitting and leads to better performance.
We also consider another hyperparameter ($l$) as the learning rate (LR) which is factor between the LR for translation to target image ($LR_o$) and LR for translation to semantic map ($LR_s$). Basically, $LR_s$ is given by 
\begin{equation}
    LR_s = \frac{LR_o}{l}.
\end{equation}
In this paper, we consider the value of $l$ as 1, 10 and 100 leading to LR Setting 1, LR Setting 10 and LR Setting 100, respectively. The illustration of different LR Settings is shown in Fig. \ref{fig:lr} for 80:20 training model.

\section{Experimental Setup}
\label{experimental_setup}
Each model is trained for a total of $100$ epochs with Learning rate 0.002 ($2e^{-3}$) for a batch size of $2$. The resolution of images is $256 \times 256$.
The remaining part of this section contains the details of GAN models and datasets used for the experiments along with the metrics used for evaluation. We use the same loss functions for Semantic to RGB training as used for Monochrome to RGB translation.

\subsection{GAN Models Used}
In order to depict the impact of the proposed injected semantic training, we use the state-of-the-art GAN models in paired scenarios, including CycleGAN and CUT models. In paired scenario, the images in different domains are registered. The details of these models are described in below:
\begin{figure*}[!t]
    \centering
    \includegraphics[width=0.8\textwidth]{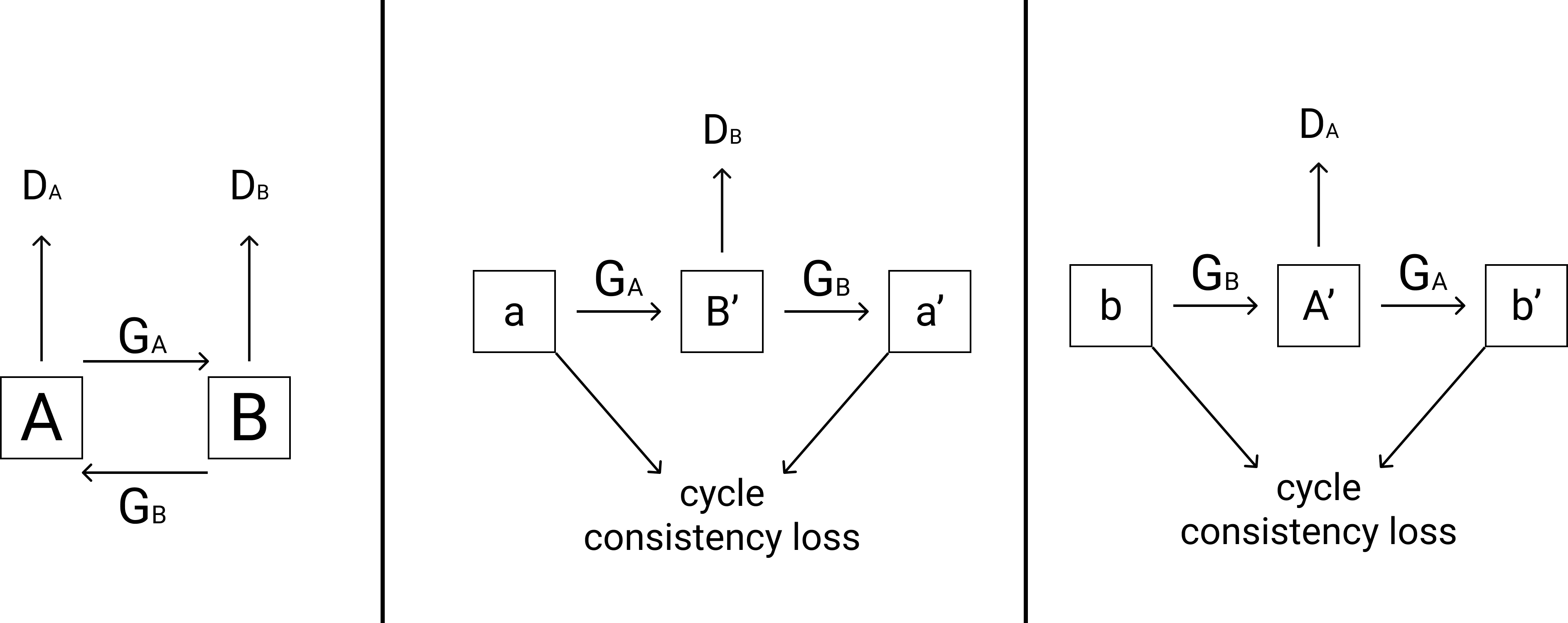}
    \caption{The illustration of CycleGAN model \cite{cyclegan}. The transformation from domain $A$ to domain $B$ is carried out using generator ($G_A$) and the transformation from domain $B$ to domain $A$ is carried out using generator ($G_B$). Cycle consistency loss is used in both the domains to facilitate the training of CycleGAN model.}
    \label{fig:CGAN}
\end{figure*}

\subsubsection{CycleGAN Model \cite{cyclegan}:}
The CycleGAN model is used to translate the image from domain A to domain B in an unpaired manner. In order to constraint the mapping, CycleGAN uses two generators, i.e., forward generator for transformation from domain A to domain B and backward generator to transform from domain B to domain A. An illustration of CycleGAN is presented in Fig. \ref{fig:CGAN} where an input image ($a$) in domain A is transformed into image ($B'$) in domain B using generator $G_A$. Further, the generated image ($B'$) is cycled back to image ($a'$) in domain A. A cycle consistency loss is computed between original image ($a$) and cycled image ($a'$) in domain A. Similarly, image in domain B ($b$) is transformed to image ($A'$) in domain A using generator $G_B$ and cycled back to image ($b'$) in domain B using generator $G_A$. A cycle consistency loss between images $b$ and $b'$ is also computed.
CycleGAN also uses two discriminators $D_A$ and $D_B$ in domain A and domain B, respectively. These discriminators distinguish between the generated samples and real samples in the corresponding domains.
The final loss for the CycleGAN model includes the adversarial loss and cycle consistency loss in both the domains. The adversarial loss consists of generator and discriminator losses. The cycle consistency loss is computed as the reconstruction error.

\begin{figure}[!t]
    \centering
    \includegraphics[width = 0.8\textwidth]{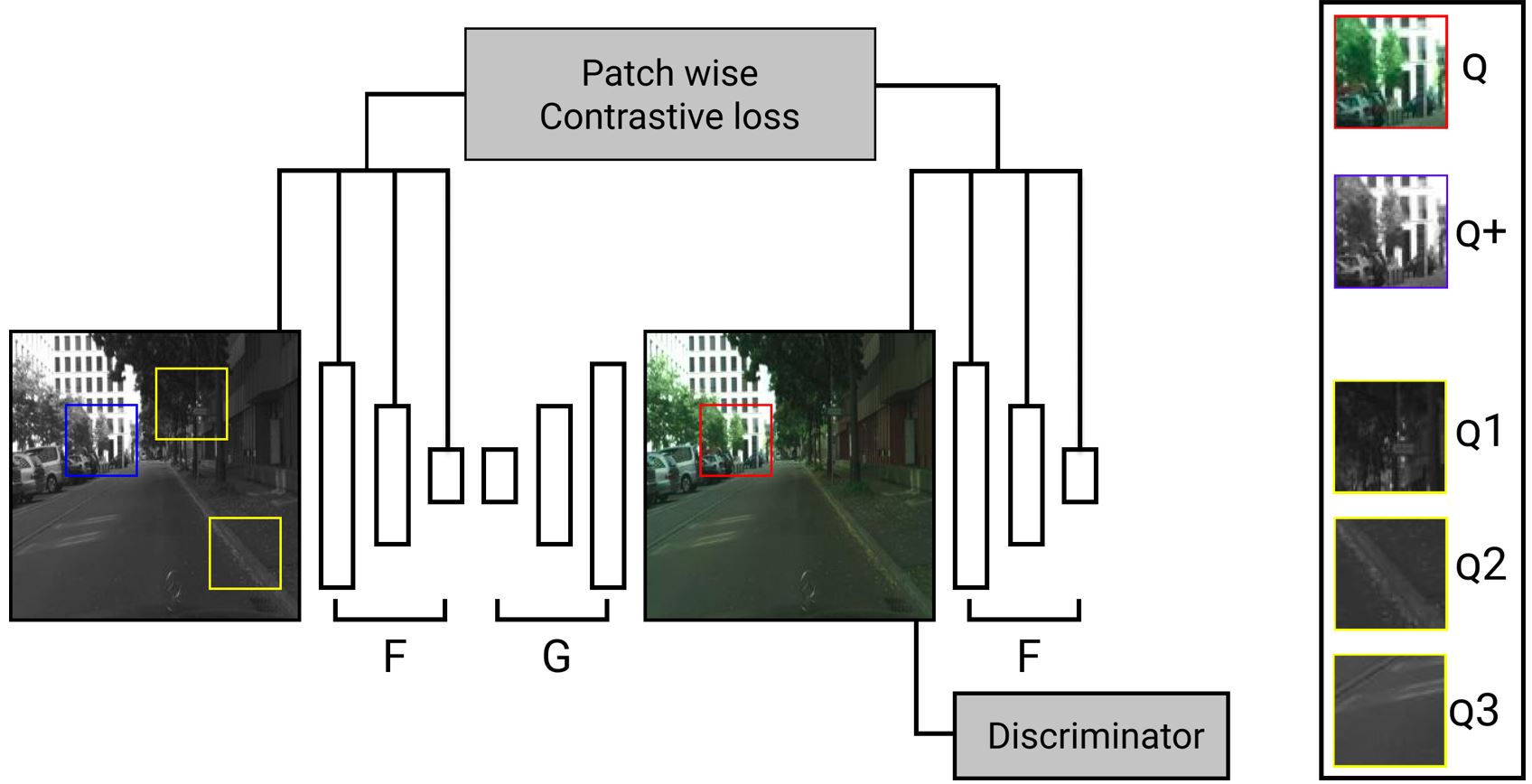}
    \caption{The illustration of CUT model \cite{cut}. The contrastive loss between the positive and negative pairs of patches is used to increase the difference between the visual appearance of positive and negative patches. Note that $F$ and $G$ are Encoder and Decoder networks, respectively.}
    \label{fig:CUT}
\end{figure}

\subsubsection{CUT Model \cite{cut}:}
The contrastive unpaired translation (CUT) model uses the contrastive loss along with the adversarial loss. Basically, it utilizes the similarity between the similar (i.e., positive) and dissimilar (i.e., negative) patches from domain A w.r.t. the patch in domain B. The patchwise contrastive loss is computed by utilizing the features of positive and negative pairs of patches. The features of patches are extracted by the encoder of the generator model.
The objective of this loss is to make the synthesized output patch closer to its corresponding input patch and apart from the other input patches. An illustration of the CUT model is presented in Fig. \ref{fig:CUT}. Consider the constructed output patch as Q, the corresponding input patch as Q+, and the other random patches as Q1, Q2, and Q3. The final loss in the CUT model is the sum of Adversarial loss of GAN and Patchwise loss in both directions, i.e., from A \small{(input domain)} to B \small{(output domain)}, and vice-versa.

\subsection{Datasets Used}
In order to demonstrate the image to image translation results, two benchmark datasets are used, namely CityScapes\footnote{\url{https://www.kaggle.com/dansbecker/cityscapes-image-pairs}} dataset and RGB-NIR\footnote{\url{http://www.cs.cmu.edu/~ILIM/projects/AA/RGBNIRStereo/}} Stereo dataset.

\begin{figure}[!t]
    \centering
    \includegraphics[width=0.45\textwidth, height=3 cm]{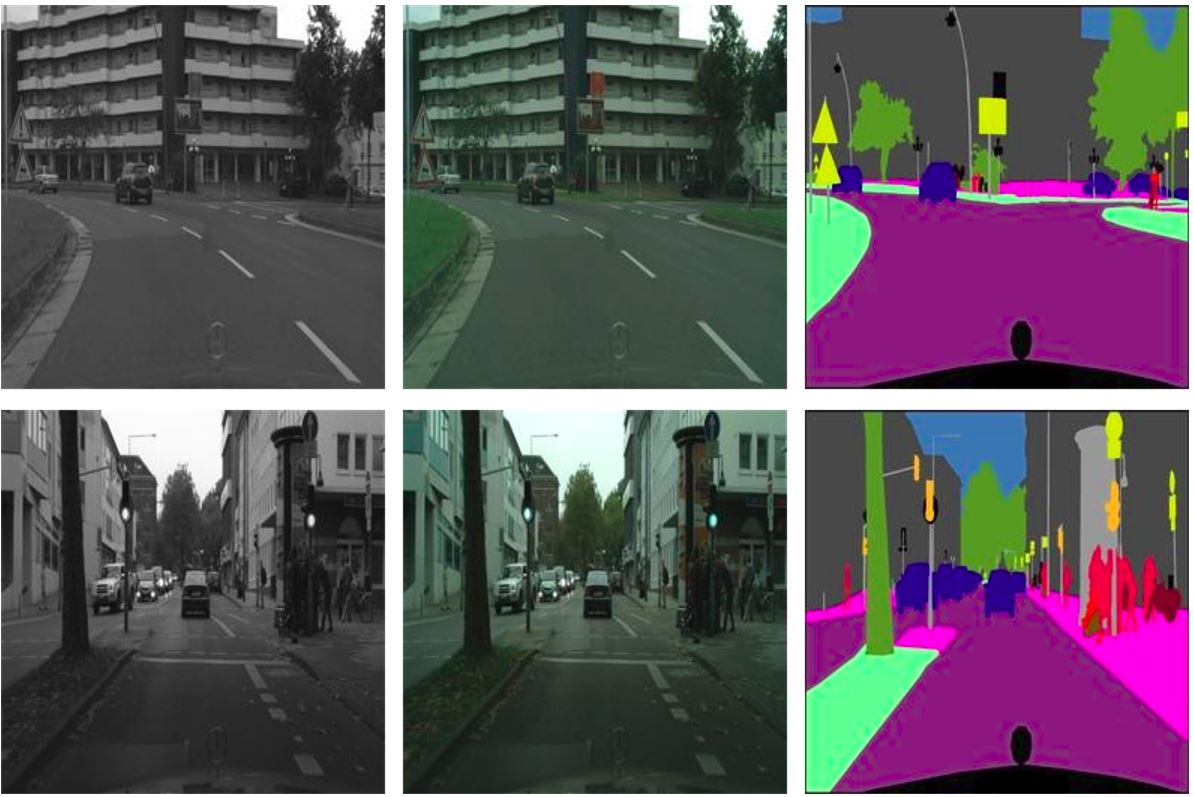}
    \hspace{1 cm}
    \includegraphics[width=0.45\textwidth, height=3 cm]{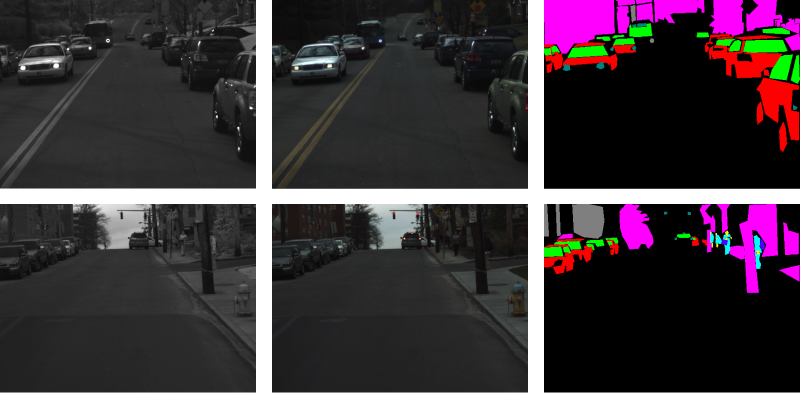}
    \caption{Example images from CityScapes ($1^{st}$ subfigure) and RGB-NIR stereo ($2^{nd}$ subfigure) datasets. In both subfigures, the input images, target images, and semantic images are shown in left, middle and right columns, respectively.}
    \label{fig:dataset}
\end{figure}

\noindent\textbf{CityScapes Dataset:}
CityScapes is a benchmark suite and large-scale dataset used to train and test techniques for semantic characterization at the pixel and instance level. CityScapes contains a wide and varied collection of stereo video sequences shot in 50 cities' streets. We consider a subset of the dataset containing a total of $2975$ training images and $500$ test images in each domain. The dataset is a set of both paired and unpaired images of resolution $256 \times 256$. The actual dataset contains only RGB and semantic images. We perform the image colorization over this dataset through image to image translation. The source images are monochrome images which are extracted from the corresponding RGB images using openCV. Thus, final dataset consists of monochrome images, corresponding RGB images and corresponding semantic images.

\noindent\textbf{RGB-NIR Stereo Dataset:}
RGB-NIR stereo is a publicly available dataset, which was prepared by collecting 13.7 hours of video data in a variety of places in and around a city using a vehicle-mounted RGB-NIR stereo system. The dataset contains materials such as lighting, glass, shiny surfaces, greenery, skin, clothing, and bags and was collected in sunny, overcast, and dark situations on college roads, highways, downtown, parks, and residential areas. This dataset is also a set of both paired and unpaired images and the resolution of images is $582 \times 429$ and contains a total of $2100$ training images and $900$ test images. This dataset contains NIR-Images (monochrome) as the source, corresponding RGB images as the target and corresponding semantic images for the injection training.

\subsection{Metrics Used}
In order to evaluate the performance of different models quantitatively, we use the Structural Similarity Index Measure (SSIM), Fréchet Inception Distance (FID), and Kernel Inception Distance (KID) metrics.

\noindent\textbf{SSIM:}
SSIM measures image quality degradation. SSIM cannot judge the better image among the two, but measures the perceptual difference between the given images. SSIM values are in the range [0, 1], where 1 denotes completely similar images and 0 denotes dissimilar images. For evaluating performance of model, we calculate the average of all SSIM values across the test dataset and report in \%.

\noindent\textbf{FID:}
FID stands for Fréchet Inception Distance, which measures the distance between feature vectors calculated for the original and synthetic (generated) images.
The score reflects how comparable the two groups of images are in terms of statistics on computer vision aspects of raw pictures calculated with the inception v3 image classification model (inception v3 module represents the given image in a vector of size $2048$). Lower scores imply that the two sets of images are more comparable with a perfect score of 0.0 for identical.

\noindent\textbf{KID:}
KID stands for Kernel Inception Distance, which is a metric similar to FID. KID also calculates the metrics using the inception v3 model representation of given images. FID measures the distance between representations, but KID measures skewness mean and variance between the vector representations. KID uses a polynomial kernel to correct the distributions and has two metrics - KID Mean and KID Variance. Lower scores imply that given two sets of images are better comparable and a perfect score of 0.0 denotes that given images are identical. 

\section{Experimental Results and Analysis}
\label{experimentalresults}
\begin{table*}[!t]
\caption{The results in terms of the SSIM and FID scores using CUT model over CityScapes dataset for different original:semantic training ratio under different learning rate (LR) setting. Note that LR Setting $l$ means the learning rate for semantic training is $1/l$ of RGB training. The best results are highlighted in bold.}
\centering
\begin{tabular}{|p{0.10\textwidth}|p{0.08\textwidth}|p{0.08\textwidth}|p{0.08\textwidth}|p{0.08\textwidth}|p{0.08\textwidth}|p{0.002\textwidth}|p{0.08\textwidth}|p{0.08\textwidth}|p{0.08\textwidth}|p{0.08\textwidth}|p{0.08\textwidth}|} 
 \hline
  LR & \multicolumn{5}{c|}{SSIM} && \multicolumn{5}{c|}{FID} \\\cline{2-12}
  Setting & 100:0 & 90:10 & 80:20 & 70:30 & 60:40 && 100:0 & 90:10 & 80:20 & 70:30 & 60:40 \\
 \hline
 1 & 93.78 & 91.91 & 94.35 & \textbf{94.53} & 93.88 && 32.50 & 36.43 & 36.59 & 38.50 & 36.30\\ 
 \hline
 10 & 93.78 & 93.22 & 93.89 & 93.81 & 94.28 && 32.50 & 32.16 & \textbf{29.75} & 34.12 & 47.25\\
 \hline
 100 & 93.78 & 94.06 & 94.14 & 94.28 & 94.49 && 32.50 & 29.76 & 33.36 & 34.51 & 38.16\\
 \hline
\end{tabular}
\label{table:cityscapes1}
\end{table*}

\begin{table*}[!t]
\caption{The results in terms of the KID Mean (KIDm) and KID Variance (KIDv) scores using CUT model over CityScapes dataset for different original:semantic training ratio under different learning rate (LR) setting. }
\centering
\begin{tabular}{|p{0.10\textwidth}|p{0.08\textwidth}|p{0.08\textwidth}|p{0.08\textwidth}|p{0.08\textwidth}|p{0.08\textwidth}|p{0.002\textwidth}|p{0.08\textwidth}|p{0.08\textwidth}|p{0.08\textwidth}|p{0.08\textwidth}|p{0.08\textwidth}|} 
\hline
LR & \multicolumn{5}{c|}{KID Mean} && \multicolumn{5}{c|}{KID Variance}  \\\cline{2-12}
Setting & 100:0 & 90:10 & 80:20 & 70:30 & 60:40 && 100:0 & 90:10 & 80:20 & 70:30 & 60:40 \\\hline
1 & 0.0105 & 0.0111 & 0.0125 & 0.0169 & 0.0134 && 0.0006 & 0.0007 & 0.0007 & 0.0007 & 0.0005 \\ 
\hline
10 & 0.0105 & 0.0091 & \textbf{0.0059} & 0.0104 & 0.0292 && 0.0006 & 0.0006 & \textbf{0.0004} & 0.0005 & 0.0011\\
\hline
100 & 0.0105 & 0.0060 & 0.0110 & 0.1285 & 0.0182 && 0.0006 & \textbf{0.0004} & 0.0007 & 0.0007 & 0.0006\\
\hline
\end{tabular}
\label{table:cityscapes2}
\end{table*}

\begin{table*}[!t]
\caption{The results in terms of the SSIM and FID scores using CUT model over RGB-NIR stereo dataset for different original:semantic training ratio under different learning rate (LR) setting. }
\centering
\begin{tabular}{|p{0.10\textwidth}|p{0.08\textwidth}|p{0.08\textwidth}|p{0.08\textwidth}|p{0.08\textwidth}|p{0.08\textwidth}|p{0.002\textwidth}|p{0.08\textwidth}|p{0.08\textwidth}|p{0.08\textwidth}|p{0.08\textwidth}|p{0.08\textwidth}|} 
\hline
LR & \multicolumn{5}{c|}{SSIM} && \multicolumn{5}{c|}{FID}  \\\cline{2-12}
Setting & 100:0 & 90:10 & 80:20 & 70:30 & 60:40 && 100:0 & 90:10 & 80:20 & 70:30 & 60:40 \\\hline
1 & 74.98 & 73.74 & \textbf{78.62} & 77.56 & 76.97 && \textbf{26.93} & 31.28 & 28.52 & 29.53 & 30.41\\ 
 \hline
10 & 74.98 & 77.40 & 75.50 & 77.27 & 75.82 && \textbf{26.93} & 29.54 & 29.37 & 29.03 & 34.32\\
 \hline
100 & 74.98 & 77.58 & 77.62 & 76.70 & 74.65 && \textbf{26.93} & 29.57 & 29.19 & 29.18 & 32.99\\
 \hline
\end{tabular}
\label{table:rgbnir1}
\end{table*}

\begin{table*}[!t]
\caption{The results in terms of the KID Mean (KIDm) and KID Variance (KIDv) scores using CUT model over RGB-NIR stereo dataset for different original:semantic training ratio under different learning rate (LR) setting. }
\centering
\begin{tabular}{|p{0.10\textwidth}|p{0.08\textwidth}|p{0.08\textwidth}|p{0.08\textwidth}|p{0.08\textwidth}|p{0.08\textwidth}|p{0.002\textwidth}|p{0.08\textwidth}|p{0.08\textwidth}|p{0.08\textwidth}|p{0.08\textwidth}|p{0.08\textwidth}|} 
\hline
LR & \multicolumn{5}{c|}{KID Mean} && \multicolumn{5}{c|}{KID Variance}\\\cline{2-12}
Setting & 100:0 & 90:10 & 80:20 & 70:30 & 60:40 && 100:0 & 90:10 & 80:20 & 70:30 & 60:40\\\hline
1 & \textbf{0.0050} & 0.0093 & 0.0070 & 0.0069 & 0.0090 && \textbf{0.0005} & 0.0008 & \textbf{0.0005} & \textbf{0.0005} & 0.0006\\ 
 \hline
10 & \textbf{0.0050} & 0.0079 & 0.0078 & 0.0070 & 0.0129 && \textbf{0.0005} & \textbf{0.0005} & 0.0007 & \textbf{0.0005} & 0.0008\\
 \hline
100 & \textbf{0.0050} & 0.0073 & 0.0072 & 0.0074 & 0.0104 && \textbf{0.0005} & 0.0006 & 0.0007 & 0.0008 & 0.0008\\
 \hline
\end{tabular}
\label{table:rgbnir2}
\end{table*}

\begin{table*}[!t]
\caption{The results in terms of the SSIM, FID, KID Mean (KIDm) and KID Variance (KIDv) scores using CycleGAN model over CityScapes and RGB-NIR stereo datasets for different original:semantic training ratio.}
\centering
\begin{tabular}{|p{0.077\textwidth}|p{0.083\textwidth}|p{0.083\textwidth}|p{0.083\textwidth}|p{0.083\textwidth}|p{0.083\textwidth}|p{0.002\textwidth}|p{0.083\textwidth}|p{0.083\textwidth}|p{0.083\textwidth}|p{0.083\textwidth}|p{0.083\textwidth}|} 
\hline
& \multicolumn{5}{c|}{CityScapes Dataset} && \multicolumn{5}{c|}{RGB-NIR Dataset}\\\cline{2-12}
Metric & 100:0 & 90:10 & 80:20 & 70:30 & 60:40 && 100:0 & 90:10 & 80:20 & 70:30 & 60:40\\\hline
SSIM & 94.02 & 94.21 & \textbf{94.89} & 94.67 & 94.17 && 76.46 & 76.95 & \textbf{79.51} & 77.79 & 74.41\\\hline
FID & 34.10 & 30.86 & \textbf{29.50} & 30.35 & 33.34 && \textbf{40.03} & 40.88 & 40.91 & 45.28 & 53.02\\\hline
KIDm & 0.0088 & 0.0071 & 0.0061 & \textbf{0.0058} & 0.0068 && \textbf{0.0127} & 0.0131 & 0.0155 & 0.0181 & 0.0239\\\hline
KIDv & 0.0005 & \textbf{0.0003} & \textbf{0.0003} & \textbf{0.0003} & \textbf{0.0003} && \textbf{0.0006} & 0.0008 & 0.0008 & 0.0007 & 0.0010\\\hline
\end{tabular}
\label{table:cyclegan}
\end{table*}

\begin{figure*}[!t]
    \centering
    \includegraphics[width=\textwidth, height=14cm]{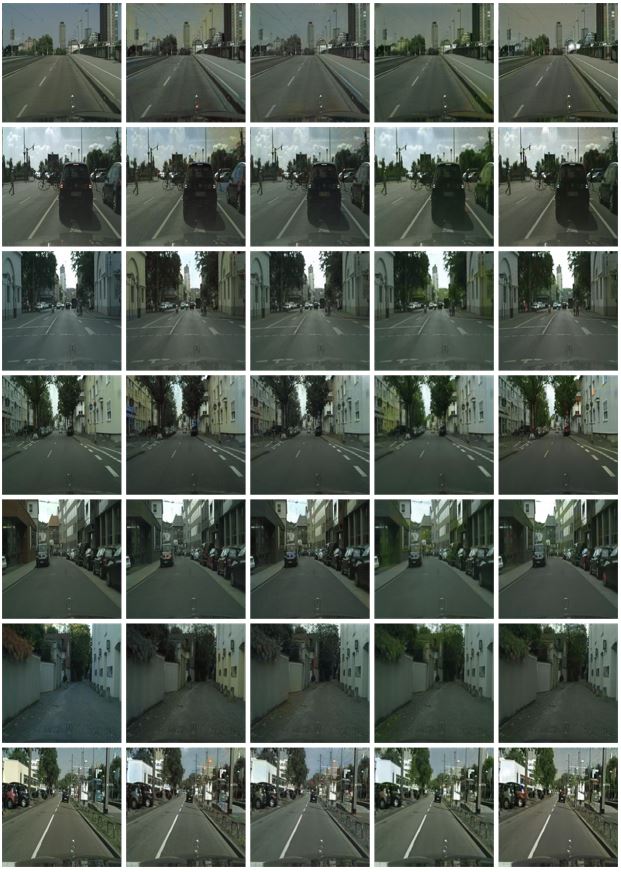}
    \caption{The qualitative results on CityScapes dataset. $1^{st}$ \textit{column}: Ground truth images, $2^{nd}$ \textit{column}: Generated images using CUT model without semantic training, $3^{rd}$ \textit{column}: Generated images using CUT model with proposed semantic training having 70:30 epochs for original:semantic, $4^{th}$ \textit{column}: Generated images using CycleGAN model without semantic training, and $5^{th}$ \textit{column}: Generated images using CycleGAN model with proposed semantic training having 80:20 epochs for original:semantic.}
    \label{r1}
\end{figure*}

By considering both the hyperparameters, i.e., original:semantic epochs ratio (i.e., 4 different ratios) and semantic LR factors (i.e., 3 different LR settings), we test 12 combinations. 
The results using CUT model are reported over CityScapes dataset in Table \ref{table:cityscapes1} in terms of the SSIM and FID and in Table \ref{table:cityscapes2} in terms of the KID Mean and KID Variance. The results using CUT model are reported over RGB-NIR stereo dataset in Table \ref{table:rgbnir1} in terms of the SSIM and FID and in Table \ref{table:rgbnir2} in terms of the KID Mean and KID Variance. In each table, the best result is highlighted in \textbf{bold}.
In each Table, $1^{st}$, $2^{nd}$ and $3^{rd}$ row results correspond to LR settings 1, 10 and 100, respectively. Whereas, the results for different original:semantic training ratios are reported in the corresponding columns. 
It is observed from these results that the performance of the proposed semantic map injected training is better than original training over CityScapes dataset in terms of the SSIM, FID and KID measures. Moreover, an improved performance is also observed on the RGB-NIR dataset by the proposed training scheme in terms of the SSIM. The training schedule with ratios 80:20 and 70:30 shows better improvement than 100:0. LR setting 1 is better suited for SSIM and LR setting 10 is better suited for other metrics.

The results in terms of the SSIM, FID and KID measures using CycleGAN model over CityScapes and RGB-NIR datasets are reported in Table \ref{table:cyclegan} for different original:semantic training ratios. In this experiment the LR setting 1 is followed. A clear improvement is gained using the proposed semantic map injection training over CityScapes dataset. However, the result over RGB-NIR is also improved in terms of the SSIM. 

The qualitative results in terms of the generated samples from CityScapes dataset are illustrated in Fig. \ref{r1} using CUT and CycleGAN models for 100:0, 70:30/80:20 training settings. The quality of the generated images is very appealing. It can be seen that semantic learning helps the model to learn the various regions in the given image. It can be also observed that the original model tends to overfit a particular color without accounting the region of that particular object. Not only the visual results, but also the metric values indicate that semantic learning helps the model to generalize better and produces better quality images. Note that the semantic map is not required at the test time.

\section{Conclusion}
\label{conclusion}
In this paper, we have proposed a semantic map injected training of GAN models for image to image translation. The semantic map training injection is performed in an interleaved manner with the original training. The proposed method ensures that the semantic map is not required at test time. The proposed semantic map injection training improves the generalization of the model and reduces overfitting. This approach is tested over CityScapes and RGB-NIR stereo benchmark datasets using CycleGAN and CUT models.  We found 30\% and 20\% of semantic injection leads to better performance as compared to the vanilla training. Both quantitative and qualitative results point out that semantic map injection training helps the model to understand the different regions of objects, and thus results in better translated images. The limitation of the proposed model is the need of semantic map at the training time. The future direction includes the extension of this work on videos and reduction of parameters leading to light weight model for real-time applications.

\section*{Acknowledgement}
This research is funded by DRDO Young Scientist Laboratory - Artificial Intelligence (DYSL-AI), Bangalore through grant no. DYSL-AI/01/2020-2021.

\bibliographystyle{splncs04}
\bibliography{References}
\end{document}